\documentclass{article}
\pdfoutput=1

\usepackage[final,nonatbib]{neurips_2024}
\usepackage{soul}

\usepackage[utf8]{inputenc} 
\usepackage[T1]{fontenc}    
\usepackage{hyperref}       
\usepackage{url}            
\usepackage{booktabs}       
\usepackage{amsfonts}       
\usepackage{nicefrac}       
\usepackage{microtype}      
\usepackage{xcolor}         

\usepackage{amsmath}
\usepackage{graphicx}
\usepackage{subcaption}
\usepackage{multirow}
\usepackage{bbding}
\usepackage{float}
\usepackage{wrapfig}
\usepackage{amssymb}

\newcommand{\secref}[1]{Sec.~\ref{#1}}
\newcommand{\figref}[1]{Fig.~\ref{#1}}
\newcommand{\tabref}[1]{Tab.~\ref{#1}}

\usepackage{colortbl}
\usepackage{booktabs}
\newcommand{\best}{\cellcolor{red!35}}
\newcommand{\sbest}{\cellcolor{orange!35}}
\newcommand{\tbest}{\cellcolor{yellow!35}}

\title{
VCR-GauS: View Consistent Depth-Normal Regularizer for Gaussian Surface Reconstruction
}

\author{%
    Hanlin Chen$^{1}$\!\!\quad Fangyin Wei$^{2}$\!\!\quad Chen Li$^{1}$\!\!\quad Tianxin Huang$^{1}$\!\!\quad Yunsong Wang$^{1}$\!\!\quad Gim Hee Lee$^1$ \\
    $^1$\, School of Computing, National University of Singapore \\ 
    $^2$\, Princeton University \\ 
    \texttt{hanlin.chen@u.nus.edu} \quad \texttt{gimhee.lee@nus.edu.sg} \\
    {\tt \href{https://hlinchen.github.io/projects/VCR-GauS/}{\textbf{https://hlinchen.github.io/projects/VCR-GauS/}}}
}

\begin{document}
\maketitle
    \begin{figure*}[h]
	\centering
         \includegraphics[width=1\linewidth]{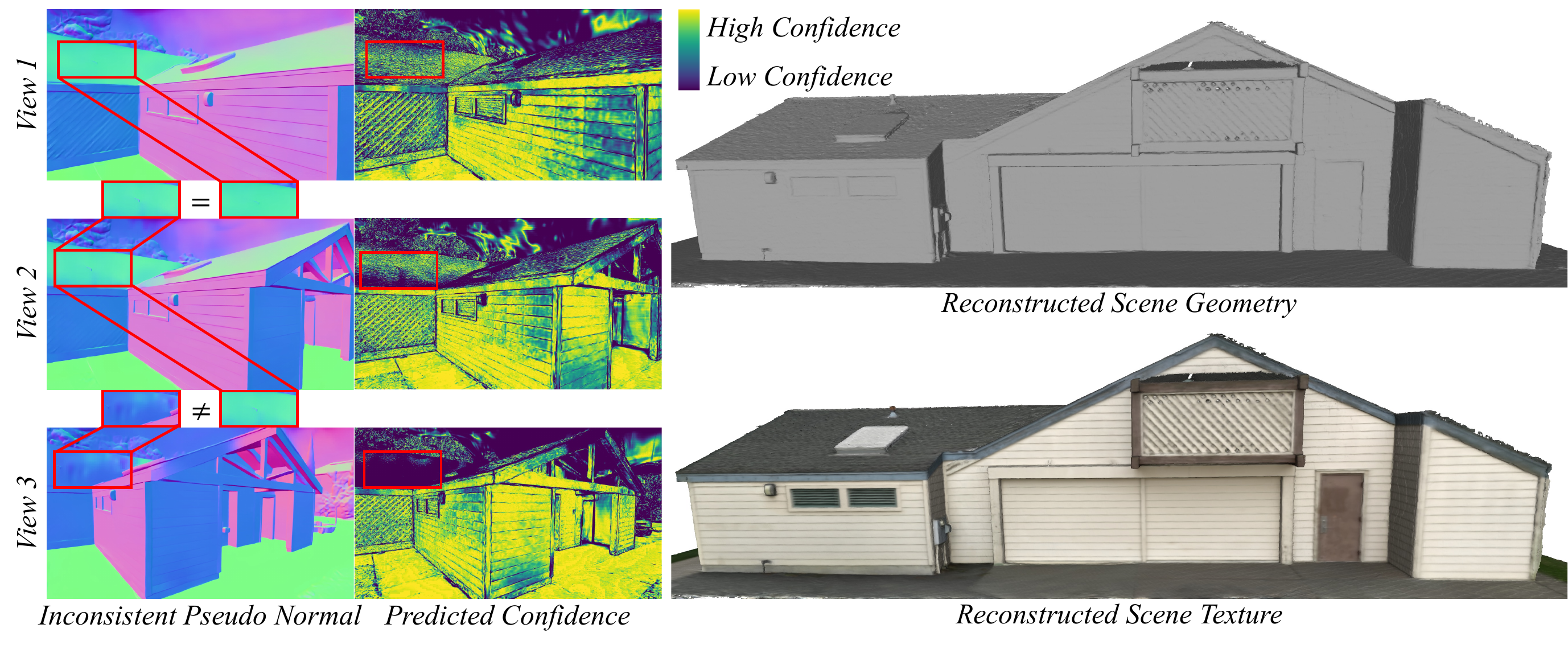}
 \caption{\textbf{View-Consistent D-Normal Regularizer.} Pseudo normals predicted from pretrained monocular normal estimators tend to be inconsistent across different views (left). Our method calculates a confidence map indicating the confidence of the pseudo normals (middle). The confidence is used to weigh the loss imposed on our proposed D-Normals. Our method achieves new state-of-the-art surface reconstruction results and rendering quality comparable with prior work.}
 \label{fig:teaser}
\end{figure*}

\begin{abstract}

Although 3D Gaussian Splatting has been widely studied because of its realistic and efficient novel-view synthesis, it is still challenging to extract a high-quality surface from the point-based representation. Previous works improve the surface by incorporating geometric priors from the off-the-shelf normal estimator. However, there are two main limitations: 1) Supervising normals rendered from 3D Gaussians effectively updates the rotation parameter but is less effective for other geometric parameters; 2) The inconsistency of predicted normal maps across multiple views may lead to severe reconstruction artifacts. In this paper, we propose a Depth-Normal regularizer that directly couples normal with other geometric parameters, leading to full updates of the geometric parameters from normal regularization. We further propose a confidence term to mitigate inconsistencies of normal predictions across multiple views. Moreover, we also introduce a densification and splitting strategy to regularize the size and distribution of 3D Gaussians for more accurate surface modeling. Compared with Gaussian-based baselines, experiments show that our approach obtains better reconstruction quality and maintains competitive appearance quality at faster training speed and 100+ FPS rendering.
\end{abstract}

\section{Introduction}
\label{sec:intro}

        \begin{figure*}[h]
    	\centering
             \includegraphics[width=1\linewidth]{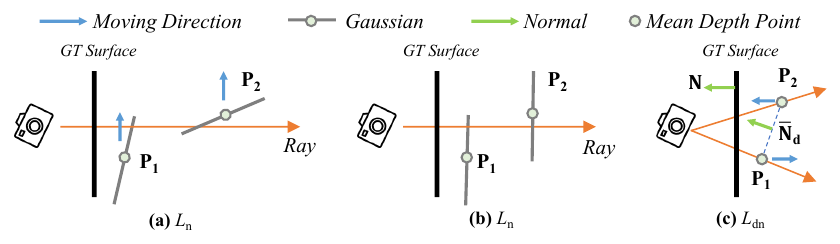}
        \caption{
        \textbf{Illustration of rendered normal supervision and the D-Normal regularizer.} (a) As a result of the back-propagation through alpha-blending via Eq.~\ref{eq:gaussian_dist}, rendered normal supervision $\mathcal{L}_{\text {n}}$ moves Gaussians closer to ($\textbf{P}_1$) or away from ($\textbf{P}_2$) the intersecting ray. 
        When the normal of a Gaussian is closer to the GT surface normal, the supervision pushes this Gaussian ($\textbf{P}_1$) towards the ray to increase its weight in the rendering equation, and vice-versa ($\textbf{P}_2$). 
        (b) Such movement of Gaussians 
        stops when the rendered normal loss $\mathcal{L}_{\text {n}}$ is equal to zero. In either case ((a) or (b)), the rendered normal loss cannot move Gaussian towards the surface. In contrast, (c) the D-Normal regularizer $\mathcal{L}_{\text {dn}}$ can move Gaussians towards or away from GT surface. $\textbf{P}_1$ and $\textbf{P}_2$ are the 3D positions corresponding to the mean depth of two neighboring pixels (rays) via Eq.~\ref{eq:depth}. The D-Normal $\bar{\textbf{N}}_d$ is derived from $\textbf{P}_1$ and $\textbf{P}_2$ in Eq.~\ref{eq:normal_depth}. $\mathcal{L}_{\text {dn}}$ encourages $\bar{\textbf{N}}_d$ to align with the ground truth normal $\textbf{N}$, resulting in Gaussians moving towards or away from the surface.}
        \label{fig:illustration_n_dn}
        \end{figure*}

Multi-view stereo (MVS) is a long-standing problem that aims to create 3D surfaces of an object or scene captured from multiple viewpoints~\cite{de1999poxels, broadhurst2001probabilistic, kutulakos2000theory, seitz1999photorealistic}. This technique has applications in robotics, graphics, virtual reality, \textit{etc}. Recently, rendering methods~\cite{wang2021neus,yu2022monosdf,li2023neuralangelo,guo2022incremental} have enhanced the quality of reconstructions. These approaches which are often based on implicit neural representations 
require extensive training time. For instance, Neuralangelo~\cite{li2023neuralangelo} uses hash encoding~\cite{muller2022instant} for creating high-fidelity surfaces but requires 128 GPU hours for a single scene. On the other hand, the novel 3D Gaussian Spatting method~\cite{kerbl3Dgaussians} employs 3D Gaussians to render complex scenes photorealistically in real-time, offering a more efficient alternative. Consequently, many recent works attempted the utilization of Gaussian Splatting for surface reconstruction~\cite{chen2023neusg,guedon2023sugar,tang2023dreamgaussian,Huang2DGS2024}. Although they achieve success in object-level reconstruction, it is still challenging to extract a high-quality surface for large scenes. Previous works~\cite{yu2022monosdf} improve the surface for scene-level reconstruction by incorporating geometric priors from the off-the-shelf normal estimator. However, there are two main limitations for Gaussian-based reconstruction: 1) Supervising normals rendered from 3D Gaussians effectively updates the rotation parameter but is less effective for other geometric parameters; 2) The predicted normal maps are inconsistent across multiple views, which may lead to severe reconstruction artifacts.


In this paper, we introduce a novel view-consistent Depth-Normal (D-Normal) regularizer to alleviate the above-mentioned limitations. 
As illustrated in \figref{fig:illustration_n_dn}, we notice that the supervision of the Gaussian normals can effectively update its rotations but is less effective for affecting its positions. 
Consequently, the supervision of Gaussian normals is not as effective as NeuS-based methods~\cite{wang2021neus,yu2022monosdf,li2023neuralangelo} whose normal is the gradient of the signed distance function (SDF) that is directly related to the position in 3D space. To solve this issue, we are inspired by the depth and normal estimation~\cite{bae2024dsine,yin2019enforcing} to introduce a D-Normal formulation, where the normal is derived from the gradient of rendered depth instead of directly blended from 3D Gaussians. Unlike existing works that obtain depth from the center position of 3D Gaussians, we compute the depth as the intersection of the ray and the compressed Gaussians. Specifically, we first make the Gaussians suitable for 3D reconstruction by applying a scale regularization similar to NeuSG~\cite{chen2023neusg} to compress the 3D Gaussian ellipsoids into a plane. Subsequently, the computation of the depth can be simplified to the intersection between a ray and a plane. As a result, our novel parametrization of the depth allows effective full supervision of the Gaussian geometric parameters by any data-driven monocular normal estimator.

To mitigate the inconsistent normal predictions across views, we further propose an uncertainty-aware normal regularizer as shown in \figref{fig:teaser}. Particularly, we introduce a confidence term for each normal prediction. A high confidence means low uncertainty leading to enhancement of the normal regularization, and vice-versa. Typically, the predicted normal maps from different views are combined to assess the uncertainty of a specific view. However, it is challenging to find correspondence across different views. We circumvent this issue by using the rendered normal learned from multi-view normal priors since we notice that it represents an average of normal priors across views. Furthermore, the confidence term is computed as the cosine distance between the rendered and predicted normals. Although the normal supervision has made the normals more accurate, there is still a minor error leading to depth error arising from the remnant large Gaussians. We thus devise a new densification that splits large Gaussians into smaller ones to represent the surface better. Finally, we incorporate a new splitting strategy to alleviate the surface bumps caused by densification.
Experiments show that our approach outperforms Gaussian-based baselines in terms of both reconstruction quality and rendering speed.

Our \textbf{main contributions} are summarized below:
\begin{itemize}
    \item 
    We formulate a novel multi-view D-Normal regularizer that enables full optimization of the Gaussian geometric parameters to achieve better surface reconstruction.
    \item 
    We further design a confidence term to weigh our D-Normal regularizer to mitigate inconsistencies of normal predictions across multiple views.
    \item
    We introduce a new densification and splitting strategy to alleviate depth error towards more accurate surface modeling.
    \item Our method outperforms prior work in terms of reconstruction accuracy and running efficiency on the benchmarking Tank and Temples, Replica, MipNeRF360, and DTU datasets.
\end{itemize}

\section{Related Work}

\noindent \textbf{Novel View Synthesis.}
    The pursuit of novel view synthesis began with Soft3D~\cite{penner2017soft}, which integrated deep learning and volumetric ray-marching to form a continuous, differentiable density field for geometry representation~\cite{henzler2019escaping, sitzmann2019deepvoxels}. While effective, this approach was computationally expensive. Neural Radiance Fields (NeRF)~\cite{nerf} improved render quality with importance sampling and positional encoding, but the deep neural networks slowed down processing. Subsequent methods aimed to optimize both quality and speed. Techniques like position encoding and band-limited coordinate networks are combined with neural radiance fields for pre-filtered scene representation~\cite{barron2021mip, barron2022mip, lindell2022bacon}. Innovations to speed up rendering included leveraging spatial data structures and adjusting MLP size~\cite{chen2022tensorf, fridovich2022plenoxels, garbin2021fastnerf, hedman2021baking, reiser2021kilonerf, takikawa2021neural}. Notable examples are InstantNGP~\cite{muller2022instant}, which uses a hash grid and a reduced MLP for faster computation, and Plenoxels~\cite{fridovich2022plenoxels}, which employs a sparse voxel grid to eliminate neural networks entirely. Both use Spherical Harmonics to enhance rendering. Despite these advancements, challenges remain in representing empty space and maintaining image quality with structured grids and extensive sampling. Recently, 3D Gaussian Splatting (3DGS)~\cite{kerbl3Dgaussians} has addressed these issues with unstructured and GPU-optimized splatting, achieving faster and higher-quality rendering without neural components. In this work, we utilize the advantage of Gaussian Splatting to perform surface reconstruction and incorporate normal priors to guide the reconstruction, especially for large indoor and outdoor scenes.

\noindent \textbf{Multi-View Surface Reconstruction.}
    Surface reconstruction is key in 3D vision. Traditional MVS methods~\cite{bleyer2011patchmatch, de1999poxels, broadhurst2001probabilistic, kutulakos2000theory, schonberger2016pixelwise, seitz1999photorealistic, seitz2006comparison} use feature matching for depth~\cite{bleyer2011patchmatch, schonberger2016pixelwise} or voxel-based shapes~\cite{de1999poxels, broadhurst2001probabilistic, kutulakos2000theory, seitz1999photorealistic, tulsiani2017multi}. Depth-based methods combine depth maps into point clouds, while volumetric methods estimate occupancy and color in voxel grids~\cite{de1999poxels, broadhurst2001probabilistic, liu2020dist}. However, the finite resolution of voxel grids limits precision. Learning-based MVS modifies traditional steps such as feature matching~\cite{luo2016efficient, ummenhofer2017demon, zagoruyko2015learning}, depth integration~\cite{riegler2017octnetfusion}, or depth inference from images~\cite{huang2018deepmvs, yao2018mvsnet, zhang2020visibility, yao2019recurrent, yu2020fast}. Further advancements~\cite{wang2021neus,yariv2021volume} integrated implicit surfaces with volume rendering, achieving detailed surface reconstructions from RGB images. These methods have been extended to large-scale reconstructions via additional regularization~\cite{yu2022monosdf,li2023neuralangelo}. Despite these impressive developments, efficient large-scale scene reconstruction remains a challenge. For example, Neuralangelo~\cite{li2023neuralangelo} requires 128 GPU hours for reconstructing a single scene from the Tanks and Temples Dataset~\cite{knapitsch2017tanks}. To accelerate the reconstruction process, some works~\cite{guedon2023sugar,Huang2DGS2024} introduce the 3D Gaussian splitting technique. However, these works still fail in large-scale reconstructions. In this work, we focus on introducing normal regularization for large-scale reconstructions.
    
    \noindent \textbf{3D Gaussian Splatting.}
    Since 3DGS~\cite{kerbl3Dgaussians} was introduced, it has been rapidly extended to surface reconstruction. We highlight the distinctions between our method and concurrent works SuGaR~\cite{guedon2023sugar}, 2DGS~\cite{Huang2DGS2024}, NeuSG~\cite{chen2023neusg}, and DN-Splatter~\cite{turkulainen2024dnsplatter}. In contrast to SuGaR and 2DGS with unsatisfactory performance on large-scale scenes, our method focuses on introducing normal regularization to improve large-scale reconstructions. 2DGS obtaining 2D Gaussian primitives by setting the last entry of scaling factors to zero which is hard to optimize by original Gaussian Splatting technique as noted in ~\cite{zwicker2004perspective,Huang2DGS2024}, while our method utilizes scale regularization to flatten 3D Gaussians which are easier to optimize. NeuSG utilizes both 3D Gaussian splitting and neural implicit rendering jointly and extracts the surface from an SDF network, while our approach is faster and conceptually simpler by leveraging only Gaussian splatting for surface approximation. Although normal prior is also used for indoor scenes, DN-Splatter may show severe reconstruction artifacts due to their normal supervision can only update the rotation parameters and  
    normal maps inconsistencies across multiple views. Moreover, we do not use the ground truth depth for supervision utilized by DN-Splatter. In comparison, our work is designed to solve both limitations.

    \begin{figure*}[t]
	\centering
	\includegraphics[width=\linewidth]{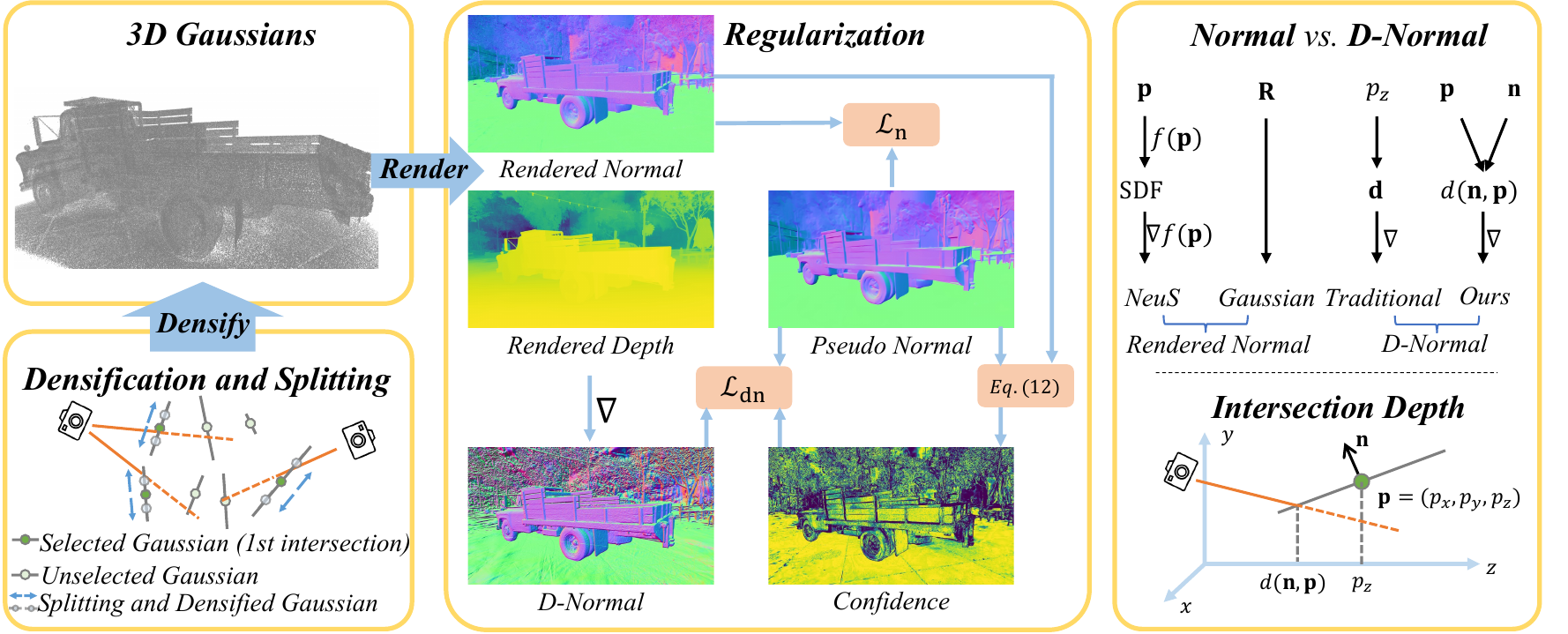}
     \caption{\label{fig:framework}\textbf{Overview of our VCR-GauS.} During densification and splitting, our method only keeps the Gaussians at the first intersections and splits large Gaussians into smaller ones along the major principle axis. The rendered normals are supervised with pseudo normals predicted from a pretrained monocular normal estimator in $\mathcal{L}_{\text{n}}$. We further calculate an uncertainty map based on the discrepancies between the rendered and pseudo normals (\textit{cf.} Eq.~\ref{eq:w}) to weigh the loss $\mathcal{L}_{\text {dn}}$ between pseudo normals and D-Normals derived from the rendered depth maps. We compare different approaches for normal calculation (Top Right) and show our intersection depth (Bottom Right).
     }
\end{figure*}

\section{Our Method}
    \label{sec:method}
    Our proposed view-consistent D-Normal regularizer efficiently reconstructs complete and detailed surfaces of scenes from multi-view images. \secref{sec:preliminary} provides an overview of 3D Gaussian Splatting~\cite{kerbl3Dgaussians}. Our normal and depth formulation of 3D Gaussians is detailed in \secref{sec:property}. \secref{sec:optim} introduces our proposed regularizations. The densification and splitting of the Gaussian is described in \secref{sec:densify_split}. \figref{fig:framework} depicts our whole framework.
    
    \subsection{Preliminaries: 3D Gaussian Splatting}
        \label{sec:preliminary}
        3D Gaussian Splatting~\cite{kerbl3Dgaussians} is an explicit 3D scene representation with 3D Gaussians. Each Gaussian is defined by a covariance matrix $\boldsymbol{\Sigma}$ and a center point $\mathbf{p} \in \mathbb{R}^3$ which is the mean of the Gaussian. The 3D Gaussian distribution can be represented as:
        \begin{equation}
            \label{eq:gaussian_dist}
            G(\mathbf{x}) = \exp{\{-\frac{1}{2}(\mathbf{x}-\mathbf{p})^\top \boldsymbol{\Sigma}^{-1}(\mathbf{x}-\mathbf{p})\}}.
        \end{equation}
        To maintain positive semi-definiteness during optimization, the covariance matrix $\boldsymbol{\Sigma}$ is expressed as the product of a scaling matrix $\mathbf{S}$ and a rotation matrix $\mathbf{R}$:
        \begin{equation}
            \boldsymbol{\Sigma} = \mathbf{R} \mathbf{S} \mathbf{S}^\top \mathbf{R}^\top,
        \end{equation}
        where $\mathbf{S}$ is a diagonal matrix, stored by a scaling factor $\mathbf{s} \in \mathbb{R}^3$, and the rotation matrix $\mathbf{R}$ is represented by a quaternion $\mathbf{r} \in \mathbb{R}^4$.
    
        For novel view rendering, the splatting technique~\cite{yifan2019differentiable} is applied to the Gaussians on the camera planes. Using the viewing transform matrix $\mathbf{W}$ and the Jacobian of the affine approximation of the projective transformation $\mathbf{J}$~\cite{zwicker2001surface}, the transformed covariance matrix $\boldsymbol{\Sigma}'$ can be determined as:
        \begin{equation}
            \boldsymbol{\Sigma}' = \mathbf{J} \mathbf{W} \boldsymbol{\Sigma} \mathbf{W}^\top \mathbf{J}^\top.
        \end{equation}
        A 3D Gaussian is defined by its position $\mathbf{p}$, quaternion $\mathbf{r}$, scaling factor $\mathbf{s}$, opacity $o \in \mathbb{R}$, and color represented with spherical harmonics coefficients $\mathbf{H} \in \mathbb{R}^k$. For a given pixel, the combined color and opacity from multiple Gaussians are weighted by Eq.~\ref{eq:gaussian_dist}. The color blending for overlapping points is:
        \begin{equation}
            \hat{\mathbf{C}} = \sum_{i \in M} \mathbf{c}_i \alpha_i \prod_{j=1}^{i-1} (1 - \alpha_j),
        \end{equation}
        where $\mathbf{c}_i$ and $\alpha_i = o_i G(\textbf{x}_i)$ denote the color and density of a point, respectively.
    
    \subsection{Geometric Properties}
        \label{sec:property}
        To reconstruct the 3D surface, we introduce two geometric properties: normal and depth of a Gaussian, which are used to render the corresponding normal map and depth map for regularization.
        
        \noindent \textbf{Normal Vector.}
            Following NeuSG~\cite{chen2023neusg}, the normal of the Gaussian can be represented as the direction of the minimized scaling factor. The normal in the world coordinate system is defined as:
                \begin{equation}
                    \mathbf{n}=\mathbf{R}[k, :] \in \mathbb{R}^3, k=\operatorname{argmin}\left(\left[s_1, s_2, s_3\right]\right),
                \end{equation}
             The normal $\textbf{n}$ and position \textbf{p} are transformed into the camera coordination system with the camera extrinsic matrix, which we subsequently take as the default unless otherwise stated.
            
        \noindent \textbf{Intersection Depth.}
            \label{sec:intersected_depth}
            The existing work~\cite{tang2023dreamgaussian} obtains the depth from the center position $\textbf{p}=(p_x, p_y, p_z)$ of each Gaussian in the camera coordinate system. However, this formulation is inaccurate and results in the depth from each Gaussian center being unrelated to its normal $\textbf{n}$ during optimization. A more reasonable depth calculation is to compute the depth of the intersection between the Gaussian and the ray emitted from the camera center. To simplify the computation of intersection and represent the surface, we incorporate a scale regularization loss $\mathcal{L}_{\text {scale}}$ from NeuSG~\cite{chen2023neusg} to squeeze the 3D Gaussian ellipsoids into highly flat shapes. This loss constrains the minimum component of the scaling factor $\mathbf{s} = (s_1, s_2, s_3)^\top \in \mathbb{R}^3$ for each Gaussian towards zero:
            \begin{equation}
                \label{eq:min_scaling}
                \mathcal{L}_{\text{s}} =  \|\min(s_1,s_2,s_3)\|_1.
            \end{equation}
            This process effectively flattens the 3D Gaussian towards a planar shape which we represent by $(\textbf{p}, \textbf{n})$. As a result, any point $\textbf{o}_p$ on the plane follows the incidence equation given by: $\textbf{n} \cdot (\textbf{o}_p -\textbf{p}) = 0$. We further denote any point $\textbf{o}_l$ on a ray that passes through the origin in 3D space as $\textbf{o}_l = \textbf{r}t$, where $t \in \mathbb{R}$ is the distance from the point to the origin along the ray. We set $\textbf{o}_l = \textbf{o}_p$ at the intersection of the ray with the plane, which we can then solve for the depth of the intersection along the $z$-axis as:
            \begin{equation}
                d(\mathbf{n}, \mathbf{p})
                = \mathbf{r}_z * (\mathbf{n} \cdot \mathbf{p}) / (\mathbf{n} \cdot \mathbf{r}),
                \label{eq:intersectDepth}
            \end{equation}
            where $\mathbf{r}_z$ is the z-value of the ray direction $\textbf{r}$. From the equation, we can see the intersection depth is related to both the position $\textbf{p}$ and the normal $\textbf{n}$ of the Gaussian. This not only offers more accurate depth calculation but also enables the D-Normal regularization to backpropagate its loss to all different Gaussian parameters.
            
    \subsection{View-Consistent D-Normal Regularization}
        \label{sec:optim}
        We first introduce our D-Normal formulation to allow the full optimization of the Gaussian geometric parameters. We further propose a confidence term to relieve the constraint from uncertain predictions and strengthen it from certain ones to avoid the wrong guidance from the inconsistent normal priors from a pretrained monocular model across multiple views.
        
        \noindent \textbf{D-Normal Regularizer.}
            \label{sec:normal}
            To improve the reconstruction quality, we utilize a normal prior $\textbf{N}$ predicted from a pretrained monocular deep neural network~\cite{bae2024dsine} to supervise the rendered normal map $\hat{\mathbf{N}}$ with L1 and cosine losses:
            \begin{align}
                \label{eq:ln}
                \mathcal{L}_{\text{n}} &= \|\hat{\mathbf{N}} - \mathbf{N}\|_1 + (1 - \hat{\mathbf{N}} \cdot \mathbf{N}), \\
                \text{where } \hat{\mathbf{N}} &= {\sum_{i \in M} \mathbf{n}_i \alpha_i \prod_{j=1}^{i-1} (1 - \alpha_j)} / {\sum_{i \in M} \alpha_i \prod_{j=1}^{i-1} (1 - \alpha_j)}.
            \end{align}

            However, normal regularization alone is insufficient for surface reconstruction as compared with NeuS-based methods. 
            There are two main reasons for this: 1) Updating the position of a Gaussian using \( G(\textbf{x}) \) only moves it closer to or farther from the intersecting ray, as shown in \figref{fig:illustration_n_dn} (a) (the mathematical proof is provided in \ref{sec:proof} of the supplemental material); 2) NeuS-based methods calculate normals as gradients of the SDF function from the input position \(\textbf{p}\) and therefore normal regularization effectively influences position updates. However, since the normal is only related to the rotation of the Gaussian in 3DGS, supervising the rendered normals does not efficiently update positions as shown in \figref{fig:framework}.
            To solve this problem and inspired by normal and depth estimation~\cite{bae2024dsine,yin2019enforcing}, we propose a new depth-normal formulation. First, we render the depth map by weighted summing the depths: 
            \begin{equation}
                \hat{D} = {\sum_{i \in M} d_i \alpha_i \prod_{j=1}^{i-1} (1 - \alpha_j)} / {\sum_{i \in M} \alpha_i \prod_{j=1}^{i-1} (1 - \alpha_j)},
                \label{eq:depth}
            \end{equation}
            where $d_i$ is the intersection depth from Eq.~\ref{eq:intersectDepth}. Subsequently, we convert the rendered depth $\hat{D}$ to a D-Normal $\bar{\textbf{N}}_d$ and use the predicted normal $\textbf{N}$ by a pretrained model to supervise the depth via the D-Normal $\bar{\textbf{N}}_d$. In particular, the D-Normal is computed by back-projecting the depth map into point clouds $\{\textbf{d}_k(\textbf{n}, \textbf{p})\}$ with the camera intrinsic matrix. The D-Normal $\bar{\textbf{N}}_d$ is then computed by the cross-product with the horizontal and vertical finite differences from the neighboring points:
            \begin{equation}
                \bar{\textbf{N}}_d(\textbf{n}, \textbf{p}) = \frac{\nabla_v \mathbf{d}(\textbf{n}, \textbf{p}) \times \nabla_h \mathbf{d}(\textbf{n}, \textbf{p})}{|\nabla_v \mathbf{d}(\textbf{n}, \textbf{p}) \times \nabla_h \mathbf{d}(\textbf{n}, \textbf{p})|}.
                \label{eq:normal_depth}
            \end{equation}
            From this equation, we can see the D-Normal is a function of both the normal $\textbf{n}$ and the position $\textbf{p}$ of Gaussians. This allows the regularization on the D-Normal to optimize both normal $\textbf{n}$ and position $\textbf{p}$. The D-Normal regularization is formulated as:
            \begin{equation}
                \mathcal{L}_{\text {dn}}=\|\bar{\textbf{N}}_d-\textbf{N}\|_1 + (1 - \bar{\textbf{N}}_d \cdot \textbf{N}).
                \label{eq:loss_dn}
            \end{equation}
            
        \noindent \textbf{Confidence.}
            Although the D-Normal regularizer resolves the issue with the Gaussian position in the supervision of its normal, the normal maps predicted by a pretrained model are not always accurate. This is especially problematic when inconsistencies arise across multiple views. We thus introduce a confidence term $w$ to emphasize the regularization for high certainty areas while reducing on low certainty areas. Typically, the normals from different views are combined to assess the certainty of a specific view. However, it is challenging to find correspondence between different views. We circumvent the challenge by using the rendered normal learned from multi-view pseudo normals, which represents an average of the pseudo normals across the views. As a result, we can use the rendered normal to gauge the uncertainty of the predicted normal in the current view. Specifically, the confidence term $w$ is computed as the cosine distance between the rendered and predicted normals, \textit{i.e.}:
            \begin{equation}
                \label{eq:w}
                w = \exp \{ (\hat{\textbf{N}}_d \cdot \textbf{N} - 1) / \gamma \},
            \end{equation}
            where $\gamma$ is a hyper-parameter. Consequently, the view-consistent D-Normal regularizer is defined as:
            \begin{equation}
                \mathcal{L}_{\text {dn}} = w * (\|\bar{\textbf{N}}_d-\textbf{N}\|_1 + (1 - \bar{\textbf{N}}_d \cdot \textbf{N})).
            \end{equation}    
        The overall loss function combining these elements is:
            \begin{equation}
                \mathcal{L}_{\text{total}} = \mathcal{L}_{\text{RGB}} + \lambda_1 \mathcal{L}_{\text{s}} + \lambda_2 \mathcal{L}_{\text{n}} + \lambda_3 \mathcal{L}_{\text{dn}},
            \end{equation}
            with $\lambda_1$, $\lambda_2$ and $\lambda_3$ balancing the individual components. $\mathcal{L}_{\text{RGB}}$ includes L1 and D-SSIM losses.

    \subsection{Densification and Splitting}
        \label{sec:densify_split}
        We observe that the original densification and splitting in Gaussian Splatting causes depth error as well as surface bumps and protrusions to appear.
        To address this issue, we further propose a new densification and splitting strategy as depicted in \figref{fig:framework} (Bottom Left).
        
\begin{figure}[htbp]
            \centering
            \begin{subfigure}[b]{0.32\textwidth}
                \centering
                \includegraphics[width=\textwidth]{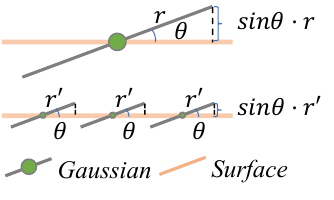}
                \caption{Large \textit{vs.} Small Gaussians}
                \label{fig:large_gau}
            \end{subfigure}
            \hfill
            \begin{subfigure}[b]{0.6\textwidth}
                \centering
                \includegraphics[width=\textwidth]{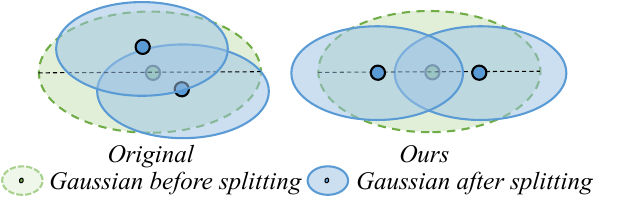}
                \caption{Splitting Strategy}
                \label{fig:split}
            \end{subfigure}
            \caption{\textbf{Illustration of the rationals behind the densification and splitting strategies.} (a) Comparison between large and small Gaussians of depth errors caused by a small normal error (in side view). (b) Comparison of the original and the proposed splitting strategies (in bird-eye view).}
            \label{fig:two_images}
        \end{figure}

        \noindent \textbf{Densification.}
            \label{sec:densify}
            Although normal supervision has made the normals more accurate, there is still a minor error $\theta$ leading to depth error arising from the remnant large Gaussians since Gaussian size is not the consideration in the original ``large position gradient" selection criteria for Gaussians to be densified. As illustrated in \figref{fig:large_gau}, a very small normal error at the edges can result in a significant depth error $ \sin\theta \cdot r $ away from the center for larger Gaussians (top of figure). Comparatively, the depth error is small for smaller Gaussians since $ r'$ becomes relatively smaller (bottom of figure). Consequently, we subdivide the larger Gaussians into smaller Gaussians to keep the depth error small. To achieve this, we first randomly sample camera views from a cuboid that encompasses the entire scene for object-centric outdoor scenes and from the training views for indoor scenes. Since we aim to densify only the surface Gaussians, we only keep the first intersected Gaussian and discard the rest for each ray emitted from the camera. Subsequently, we densify only those with a scale above a threshold $\beta$ among the collected Gaussians.

        \noindent \textbf{Splitting.}
            We notice that the Gaussians tend to protrude the ground truth surface after densification due to the clustering of many Gaussians. As also observed in Mip-Splatting~\cite{Yu2023MipSplatting}, Gaussians splitted from the same parents tend to remain clustered with relatively stable positions due to the sampling from the same Gaussian distribution. To avoid clustering, we split the old Gaussian into two new Gaussian along the axis with the largest scale instead of using the Gaussian sampling with the position of the Gaussian as mean and the 3D scale of the Gaussian as variance. The positions of the new Gaussians evenly divide the maximum scale of the old Gaussian. Other parameters of new Gaussians are obtained following the original 3DGS. This process is shown in \figref{fig:split}.
    
\section{Experiments}
    We first evaluate our method on 3D surface reconstruction in \secref{sec:rec}. We also report the rendering results in \secref{sec:render}. Additionally, we validate the effectiveness of the proposed techniques in \secref{sec:ablation}.
    
    \begin{table*}[hb]
    \centering
    \caption{\textbf{Quantitative results on the Tanks and Temples Dataset~\cite{knapitsch2017tanks}}. Reconstructions are evaluated with the official evaluation scripts and we report F1-score, average optimization time and FPS. Ours outperforms all 3DGS-based surface reconstruction methods by a large margin and performs better than neural implicit methods by a minor margin while optimizing significantly faster.}
    \begin{tabular}{@{}lccccccc}
        \toprule
     & \multicolumn{3}{c@{}}{NeuS-based} & \multicolumn{4}{c@{}}{Gaussian-based} \\ 
     & NeuS & MonoSDF & Geo-Neus & SuGaR & 3DGS & 2DGS & Ours\\ 
     \midrule
    Barn & 0.29 & 0.49 & 0.33 & 0.14 & 0.13 & \tbest 0.36 & \best 0.62\\
    Caterpillar &  \sbest 0.29 & \best 0.31 & 0.26 & 0.16 & 0.08 & 0.23 & \tbest 0.26\\
    Courthouse &  \sbest 0.17 & 0.12 & 0.12 & 0.08 & 0.09 & 0.13 & \best 0.19\\
    Ignatius &   \sbest 0.83 & \best 0.78 & \tbest 0.72 & 0.33 & 0.04 & 0.44 & 0.61\\
    Meetingroom &   \best 0.24 & \sbest 0.23 & \tbest 0.20 &  0.15 & 0.01 & 0.16 & 0.19\\
    Truck &  \sbest 0.45 & 0.42 & \tbest 0.45 &  0.26 & 0.19 & 0.26 & \best 0.52\\ 
    \midrule
    Mean &  \tbest 0.38 & \sbest 0.39 & 0.35 & 0.19 & 0.09 & 0.30 & \best 0.40\\
    Time & >24h & >24h & >24h & >1h & \best 14.3~m & \sbest 34.2~m & \tbest 53~m\\
    \midrule
    FPS & \multicolumn{3}{c@{}}{<10} & - & 159 & 68 & 145 \\
        \bottomrule
    \end{tabular}
    \label{tab:tnt}
\end{table*}

        \begin{figure*}[h]
            \centering
            \includegraphics[width=\linewidth]{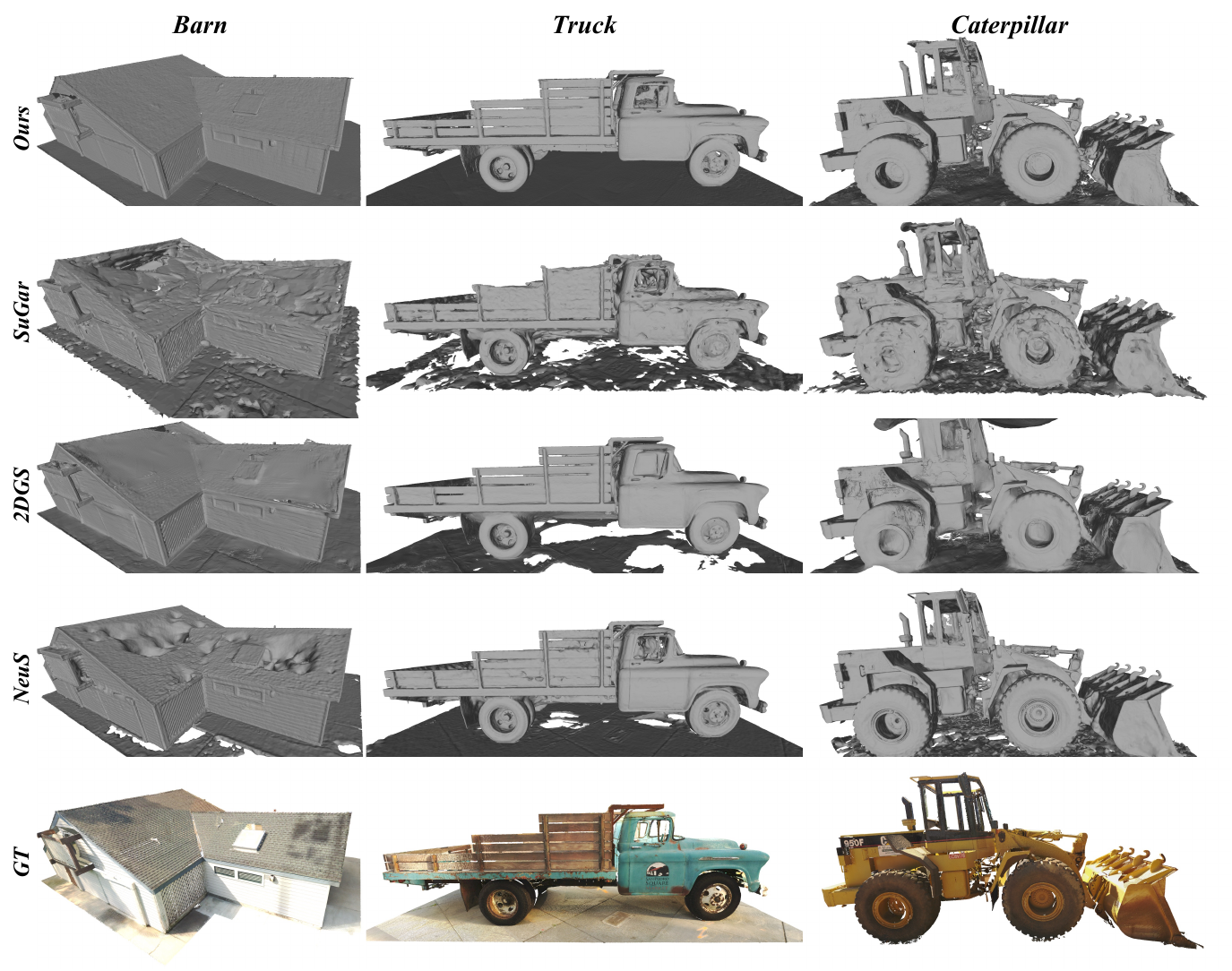} 
        \caption{\label{fig:tnt_mesh}\textbf{Qualitative comparison on TNT dataset.} From top to bottom, we show the reconstructed meshes from our method, SuGar, 2DGS, and NeuS, as well as the ground truth colored point cloud. Our method reconstructs more complete surfaces featuring smoother planar regions and finer details.}
        \end{figure*}
    \noindent \textbf{Dataset.}
        We evaluate the performance of our method on various datasets. For surface reconstruction, we evaluate on Tanks and Temples (TNT)~\cite{knapitsch2017tanks}. To further validate the effectiveness of our method, we compare with other methods on Replica~\cite{straub2019replica}. Although we focus on the large-scale reconstruction, we also report our results on DTU~\cite{dtu}, which can be seen in the supplementary. Furthermore, we evaluate the rendering results on Mip-NeRF360~\cite{barron2022mip}. For all the datasets, we use COLMAP~\cite{schoenberger2016sfm} to generate a sparse point cloud for each scene as initialization.

\begin{table}[h]
    \centering
    \caption{\textbf{Quantitative results on Mip-NeRF 360~\cite{barron2022mip}.}
    Our method achieves NVS rendering quality and speed comparable with other Gaussian-based methods.
    }
    \begin{tabular}{@{}lccccccc}
    \toprule
     & \multicolumn{3}{c@{}}{Outdoor Scene} & \multicolumn{3}{c@{}}{Indoor Scene} & \multirow{2}{*}{FPS~$\uparrow$} \\ 
    & PSNR~$\uparrow$ & SSIM~$\uparrow$ & LPIPS~$\downarrow$ & PSNR~$\uparrow$ & 
    SSIM~$\uparrow$ & LIPPS~$\downarrow$ \\
    \midrule
    NeRF & 21.46 & 0.458 & 0.515 & 26.84 &  0.790 & 0.370 & \multirow{5}{*}{<10} \\
    Deep Blending & 21.54 &0.524 & 0.364 & 26.40 & 0.844 & 0.261 \\
    Instant NGP & 22.90 & 0.566 & 0.371 & 29.15 & 0.880 & 0.216 \\
    MERF & 23.19 & 0.616 &   0.343 & 27.80 & 0.855 & 0.271 \\
    MipNeRF360 &   24.47 &  0.691 &  0.283 &   \best 31.72 & \tbest 0.917 &  \best 0.180 \\
    \hline
    \hline
    Mobile-NeRF & 21.95 & 0.470 & 0.470 & - & - & - & \multirow{2}{*}{<100} \\
    BakedSDF & 22.47 & 0.585 &  0.349 & 27.06 & 0.836 & 0.258 \\
    \hline
    \hline
    3DGS &  \best 24.64 &  \best 0.731 &  \best 0.234 &   \tbest 30.41 &  \sbest 0.920 &  \tbest 0.189 & 134 \\
    SuGaR &  22.93 & 0.629 & 0.356 & 29.43 & 0.906 & 0.225 & - \\
    2DGS &  \tbest 24.21 & \sbest 0.709 & \sbest 0.276 & 30.10 & 0.913 & 0.211 & 27 \\
    Ours & \sbest 24.31 & \tbest 0.707 &\tbest  0.280 & \sbest 30.53 & \best 0.921 &\sbest  0.184 & 128 \\
     \bottomrule
    \end{tabular}
    \vspace{-0.5cm}
\label{tab:mipnerf360}
\end{table}

    \noindent \textbf{Implementation Details.}
        Our method is built upon the open-source 3DGS code base~\cite{kerbl3Dgaussians} and the intersection depth calculation is implemented with custom CUDA kernels. $\lambda_1$, $\lambda_2$, and $\lambda_3$ are set to $1$, $0.01$, and $0.015$, respectively. The densification threshold $\beta$ is set to $0.002$. The hyperparameter $\gamma$ is set to $0.005$. We use pretrained DSINE~\cite{bae2024dsine} to predict normal maps for outdoor scenes and pretrained GeoWizard~\cite{fu2024geowizard} for indoor scenes. We also employ a semantic surface trimming approach to avoid unwanted background elements like the sky in the reconstructions for outdoor scenes, which is introduced in the supplementary material. Similar to 3DGS, we stop densification at 15k iterations and optimize all of our model parameters for 30k iterations. For mesh extraction, we adapt truncated signed distance fusion (TSDF) to fuse the rendered depth maps using Open3D~\cite{Zhou2018}.
        
    \subsection{Comparison}
            \label{sec:rec}

\begin{wraptable}{r}{0.45\textwidth}
    \vspace{-0.4cm}
    \centering
    \small
    \caption{\small 
             \textbf{Quantitative assessment on Replica \cite{straub2019replica}}. \textbf{Bold} indicates the best.
    }
    \vspace{-0.15cm}
    \label{tab:replica}
    \begin{tabular}{cccc}
    \toprule
    & Method & F1-score & Time \\
    \midrule
    \multirow{2}{*}{Implicit} & NeuS & 65.12 & \multirow{2}{*}{$>$10h} \\
    & MonoSDF & \textbf{81.64} & \\
    \midrule
    \multirow{4}{*}{Explicit} & 3DGS & 50.79 & \multirow{4}{*}{$\le$1h} \\
    & SuGar & 63.20 & \\
    & 2DGS & 64.36 & \\
    & Ours & \textbf{78.17} & \\
    \bottomrule
    \end{tabular}
    \vspace{-0.1cm}
\end{wraptable}
            \textbf{Surface Reconstruction.} As shown in~\tabref{tab:tnt}, our method outperforms SDF models (\textit{i.e.}, NeuS~\cite{wang2021neus}, 
            MonoSDF~\cite{yu2022monosdf}, and Geo-NeuS~\cite{fu2022geo}) on the TNT dataset, and reconstructs significantly better surfaces than explicit reconstruction methods (\textit{i.e.}, 3DGS \cite{kerbl3Dgaussians}, SuGaR \cite{guedon2023sugar}, and 2DGS \cite{Huang2DGS2024}). Notably, our model demonstrates exceptional efficiency, offering a reconstruction speed that is approximately 20 times faster compared to NeuS-based reconstruction methods. Compared with the concurrent work 2DGS\cite{Huang2DGS2024}, although our method is a little slower than it (about 20 minutes), it works much better (0.3 \textit{vs.} 0.4). As shown in ~\figref{fig:tnt_mesh}, our approach better recovers planar surfaces (\textit{e.g.}, roof in \textit{Barn} and ground in \textit{Truck}) as well as finer geometry details. In addition, 2DGS renders much slower than ours (68 FPS \textit{vs.} 145 FPS). On the Replica dataset shown in \tabref{tab:replica}, our method is much faster than MonoSDF (10+ hours \textit{vs.} ~50 minutes) although showing comparable performance. Compared with explicit reconstruction methods, including 3DGS, SuGaR, and 2DGS, our method achieves significantly higher F1-score for reconstruction. Although we focus on large-scale reconstruction, we also report the results on object-level reconstruction DTU~\cite{dtu} (\textit{cf.} supplementary).
        
        \noindent \textbf{Novel View Synthesis.}
            \label{sec:render}
            Our method can reconstruct 3D surfaces and provide high-quality novel view synthesis. As shown in \tabref{tab:mipnerf360}, we compare our novel view rendering results against baseline approaches on the Mip-NeRF360 dataset in this section. Remarkably, our method consistently achieves competitive novel view synthesis results compared to state-of-the-art techniques (\textit{e.g.}, Mip-NeRF360, 3DGS, \textit{etc}.) while providing geometrically accurate surface reconstruction. Furthermore, our method renders a few times faster than the concurrent work 2DGS (128 FPS \textit{vs.} 27 FPS). The reason is that our method utilizes the original and efficient Gaussian Splatting technique, while 2DGS applies a time-consuming ray-splat technique.
            
        \begin{figure*}[th]
    	\centering
             \includegraphics[width=1\linewidth]{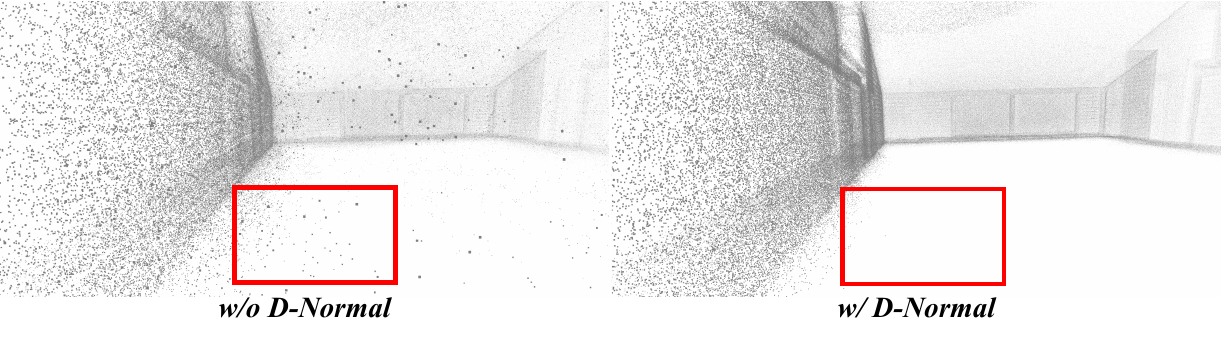}
        \caption{\textbf{Qualitative ablation for the D-Normal regularizer.}
        We plot the positions of Gaussian centers from the optimized 3D scenes. The left disables the D-Normal regularizer with only rendered normal supervision, and the right enables both D-Normal and rendered normal supervision. Compared to \textit{w/o D-Normal} that produces many noisy Gaussians floating off the surface, our proposed D-Normal regularizer effectively pushes the 3D Gaussians towards the surface and thus providing much cleaner reconstruction.
        }
        \label{fig:wodn}
        \end{figure*}
    
    \subsection{Ablation Studies}
            \label{sec:ablation}

\begin{figure*}[h]
        	\centering
        	\includegraphics[width=1\linewidth]{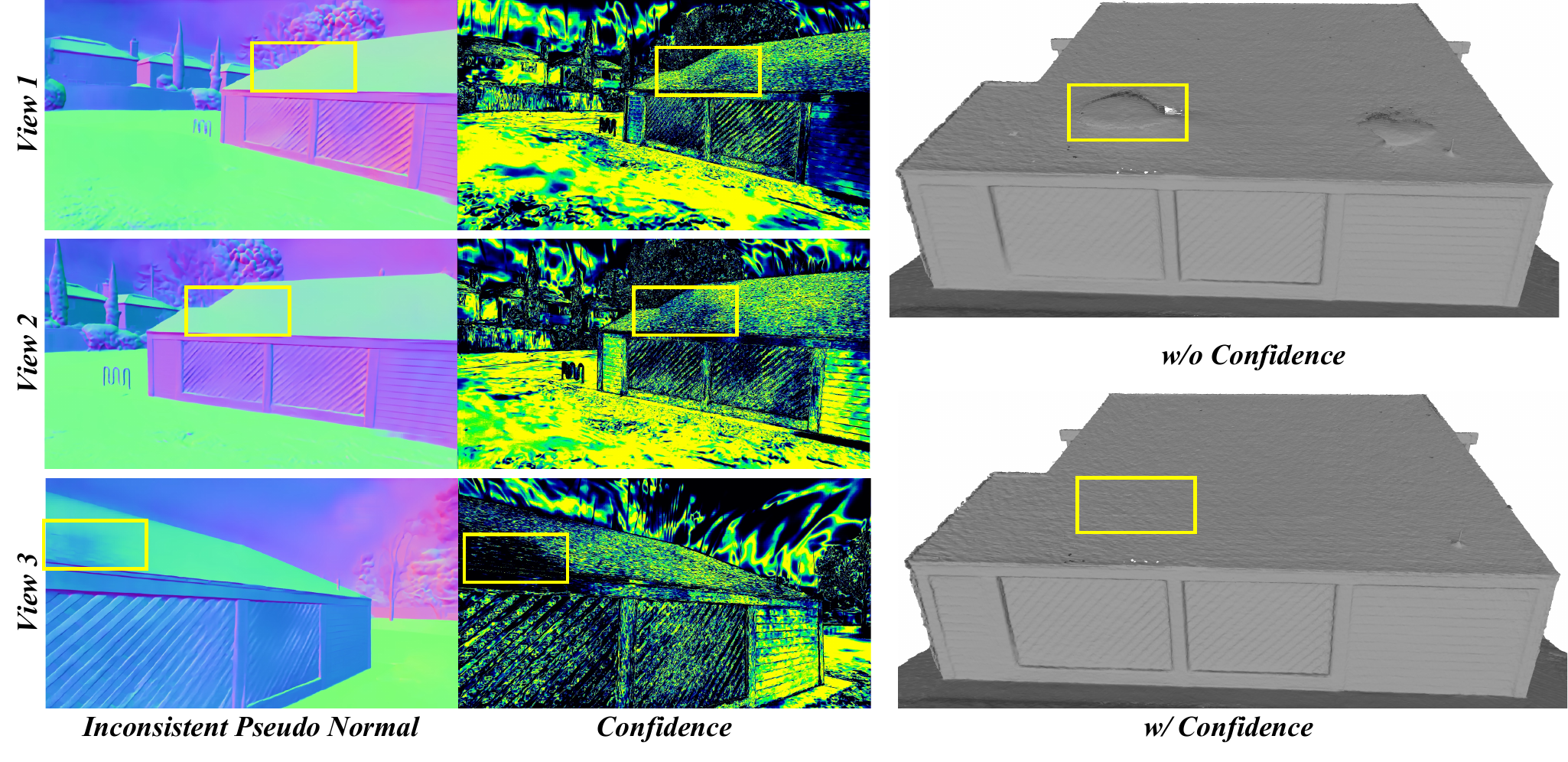}
             \caption{\label{fig:confidence}\textbf{Qualitative ablation for the confidence.}
             Without the confidence weight, the reconstructed surface shows protrusions caused by the inconsistent pseudo normal maps across different views.
             }
        \end{figure*}

\begin{wraptable}{r}{0.6\textwidth}
    \vspace{-0.4cm}
    \centering
    \small
    \caption{\textbf{Ablation on TNT~\cite{knapitsch2017tanks}.} \textbf{Bold} indicates best result.}
    \vspace{-0.15cm}
    \begin{tabular}{lccc}
    \toprule
     Ablation Item& Precision~$\uparrow$ & Recall~$\uparrow$ & F-score~$\uparrow$ \\
    \midrule
    A. w/o D-Normal & 0.27 & 0.34 & 0.30 \\
    B. w/o confidence & 0.36 & 0.37 & 0.36 \\
    C. w/o intersection depth & 0.35 & 0.37 & 0.35 \\
    D. w/o densify and split & 0.32 & 0.35 & 0.33 \\
    \midrule
    E. Full & \textbf{0.39} & \textbf{0.42} & \textbf{0.40} \\
    \bottomrule
    \end{tabular}
    \label{tab:tnt_ablation}
    \vspace{-0.3cm}
\end{wraptable}
        We verify the effectiveness of different design choices on reconstruction quality, including regularization terms, intersection depth, and densification on the TNT dataset~\cite{knapitsch2017tanks} and report the F1-score. We first examine the effect of our view-consistent D-Normal regularization. Our full model (\tabref{tab:tnt_ablation} E) provides the best performance (0.40 F1-score). The performance drops 0.10 F1-score from 0.4 to 0.3 without the D-Normal regularizer (\tabref{tab:tnt_ablation} A) while keeping rendered normal regularization. It proves that it is insufficent to supervise only the normal maps rendered from Gaussian Splatting. The visualization in \figref{fig:wodn} demonstrates that our d-normal regularization can effectively push the 3D Gaussians towards the surface. Furthermore, the result drops by 0.04 F1-score without the confidence (\tabref{tab:tnt_ablation} B) and with the D-Normal regularizer. It demonstrates that confidence can mitigate the problem of inconsistency of the predicted normal maps. From \figref{fig:confidence}, we can observe that disabling the confidence leads to an unsmooth surface. Both of these validate the effectiveness of the view-consistent D-Normal regularization. Additionally, the absence of intersection depth (\tabref{tab:tnt_ablation} C) results in poor performance. Lastly, the performance increases from 0.33 F1-score to 0.40 with our densification and split (\tabref{tab:tnt_ablation} D), proving small Gaussians represent surfaces better than large Gaussians.

    \vspace{-0.5cm}
\section{Conclusion}
In this work, we have introduced a view-consistent D-Normal regularizer for efficient, high-quality, and compact surface reconstruction. We formulate the D-Normal regularizer that directly couples normal with the other geometric parameters. This allows for the full update of all geometric parameters during normal regularization. We also propose a confidence term that weighs our D-Normal regularizer to mitigate inconsistencies of normal predictions across multiple views. Finally, we introduce a densification and splitting strategy to regularize the scales and distribution of 3D Gaussians for more precise surface modeling. Our evaluations on diverse datasets demonstrate that our method outperforms existing works in surface reconstruction.

    \noindent \textbf{Acknowledgement.} This research / project is supported by the National
Research Foundation (NRF) Singapore, under its NRF-Investigatorship Programme
(Award ID. NRF-NRFI09-0008).

\clearpage

\bibliographystyle{plain} 

\bibliography{main}


\newpage
    
\appendix

\section{Supplemental Material}


    \subsection{Semantic Surface Trimming}
        \begin{figure*}[h]
        	\centering
        	\includegraphics[width=\linewidth]{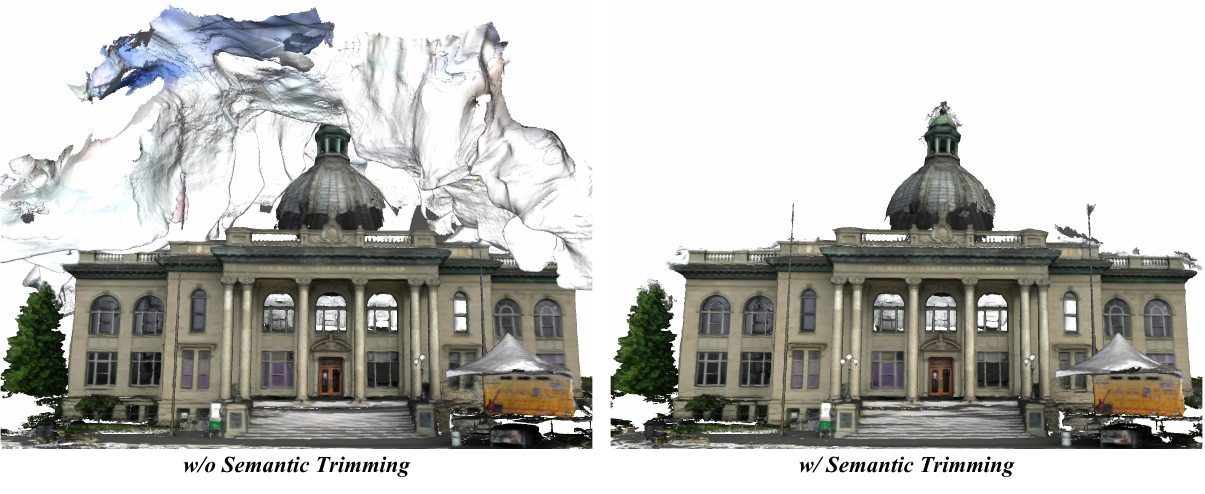}
         \caption{\textbf{Qualitative ablation for the semantic trimming.} The left is disabling the semantic trimming and the right is enabling the strategy. We can see that the proposed semantic trimming can prune the unwanted regions, 
         \textit{e.g.} the sky in the left figure.}
         \label{fig:semantic}
        \end{figure*}
        To avoid unwanted background elements like the sky in reconstructions for outdoor scenes, we employ a semantic surface trimming approach by learning a semantic field~\cite{gaussian_grouping,chen2023gnesf,wang2024gov,semantic_nerf} that leverages a pretrained semantic model~\cite{kirillov2023segany,liu2023grounding,ren2024grounded}. Specifically, we assign each Gaussian with a learnable semantic feature. Subsequently, similar to color blending, alpha blending is applied on these features to get the pixel-level semantics. We use the predicted semantic map from Grounded-SAM~\cite{ren2024grounded} with cross-entropy loss to train the learnable features. After the optimization, we render the semantic map for each view and then use the semantics to mask out the background. Although we can use the predicted semantic maps from Grounded-SAM to prune the background directly, the predicted semantic maps are not always accurate and consistent across views. Our proposed method can get a more accurate semantics with noisy pseudo labels to a certain extent, which is also observed in~\cite{semantic_nerf}.

    \subsection{Proof on Our D-Normal Regularizer}
        \label{sec:proof}

        Fig.~\ref{fig:illustration_n_dn} is to illustrate the optimization of positions of Gaussians under normal and d-normal supervisions. As we mentioned in Sec.~\ref{sec:intro}, in contrast to supervision on rendered normal maps which only updates Gaussian rotations, our D-Normal regularizer can also effectively update the positions of the Gaussians. We show the mathematical proof below.

        \vspace{2mm}
        \noindent {\textbf{Proposition 1.1:}} Supervision on rendered normal cannot effectively affect the positions of Gaussians.   

        \vspace{2mm}
        \noindent {\textbf{Proposition 1.2:}} Supervision on our D-Normal regularizer can effectively affect the positions of Gaussians.
        
        
        
        \vspace{2mm}
        \noindent \textbf{\textit{Proof:}} Without a loss of generality, we omit the summation over multiple views in our following derivations for brevity.
        Based on the loss $\mathcal{L_{\text {n}}}$ on rendered normal (\textit{cf.} Eq.~\ref{eq:ln}), the gradient of $\mathcal{L_{\text {n}}}$ with respect to position $\mathbf{p}$:
        \begin{align}
        \frac{\partial \mathcal{L_{\text {n}}}}{\partial \mathbf{p}_i} &= \frac{\mathcal{L_{\text {n}}}  } {\partial \hat{\mathbf{N}}} \cdot \frac{\partial \hat{\mathbf{N}}}{\partial \mathbf{p}_i}, \label{eq:dln_dp} \\ 
        \frac{\partial \hat{\mathbf{N}}}{\partial \mathbf{p}_i} &= \frac{\partial \hat{\mathbf{N}}}{\partial \alpha_i} \cdot \frac{\partial \alpha_i}{\partial \mathbf{p}_i} + \frac{\partial \hat{\mathbf{N}}}{\partial \mathbf{n}_i} \cdot \frac{\partial \mathbf{n}_i}{\partial \mathbf{p}_i} \nonumber \\
        &= \frac{\partial \hat{\mathbf{N}}}{\partial \alpha_i} \cdot \frac{\partial \alpha_i}{\partial G(\mathbf{x})} \cdot \frac{\partial G(\mathbf{x})}{\partial \mathbf{p}_i} \nonumber \\
        &=\frac{\partial \hat{\mathbf{N}}}{\partial \alpha_i} \cdot \frac{\partial \alpha_i}{\partial G(\mathbf{x})} \cdot [ -G(\mathbf{x}) \cdot (\mathbf{R} \mathbf{S} \mathbf{S}^\top \mathbf{R}^\top)^{-1} \cdot (\mathbf{x} - \mathbf{p}_i)] \nonumber \\
        &\approx \frac{\partial \hat{\mathbf{N}}}{\partial \alpha_i} \cdot \frac{\partial \alpha_i}{\partial G(\mathbf{x})} \cdot [ -G(\mathbf{x}) \cdot (\mathbf{x} - \mathbf{p}_i)] \label{eq:dn_dp} \\
        \nonumber 
        \end{align}
        Putting Eq.~\ref{eq:dn_dp} into Eq.~\ref{eq:dln_dp}, we get:
        \begin{align}
        \frac{\partial \mathcal{L_{\text {n}}}}{\partial \mathbf{p}_i} &\approx \frac{\mathcal{L_{\text {n}}}  } {\partial \hat{\mathbf{N}}} \cdot \frac{\partial \hat{\mathbf{N}}}{\partial \alpha_i} \cdot \frac{\partial \alpha_i}{\partial G(\mathbf{x})} \cdot [ -G(\mathbf{x}) \cdot (\mathbf{x} - \mathbf{p}_i)] \nonumber \\
        &= \beta \cdot \frac{\partial \alpha_i}{\partial G(\mathbf{x})} \cdot [ -G(\mathbf{x}) \cdot (\mathbf{x} - \mathbf{p}_i)] \nonumber \\
        &\propto (\mathbf{x} - \mathbf{p}_i), \text{where } \beta = \frac{\mathcal{L_{\text {n}}}  } {\partial \hat{\mathbf{N}}} \cdot \frac{\partial \hat{\mathbf{N}}}{\partial \alpha_i} \text{is a scalar}.
        \end{align}
        Based on the D-Normal regularization $\mathcal{L_{\text {dn}}}$ (cf. Eq.~\ref{eq:loss_dn}), the gradient of $\mathcal{L}_{\text {dn}}$ with respect to position $\mathbf{p}$:
        \begin{align}
        \frac{\partial \mathcal{L}_{\text {dn}}}{\partial \textbf{p}_i} &= \frac{\partial \mathcal{L}_{\text {dn}}}{\partial \bar{\textbf{N}}_d} \cdot \frac{\partial \bar{\textbf{N}}_d}{\partial \hat{D}} \cdot \frac{\partial \hat{D}}{\partial \textbf{p}_i}, \nonumber \\
        \frac{\partial \hat{D}}{\partial \textbf{p}_i} &= \frac{\partial \hat{D}}{\partial \alpha_i} \cdot \frac{\partial \alpha_i}{\partial \textbf{p}_i} + \frac{\partial \hat{D}}{\partial d_i} \cdot \frac{\partial d_i}{\partial \textbf{p}_i} \nonumber \\
        &= \frac{\partial \hat{D}}{\partial \alpha_i} \cdot \frac{\partial \alpha_i}{\partial G(\textbf{x})} \cdot \frac{\partial G(\textbf{x})}{\partial \textbf{p}_i} + \frac{\partial \hat{D}}{\partial d_i} \cdot r_z \cdot \frac{\mathbf{n} }  {\mathbf{n} \cdot \mathbf{r}} \label{eq:dd_dp}.
        \end{align}
        We can deduce the following from Eq.~\ref{eq:dln_dp} and Eq.~\ref{eq:dd_dp}:

        \vspace{2mm}
        \noindent \textbf{{Case 1:}}
        From Eq.~\ref{eq:dln_dp}, we can see that the gradient-update $\frac{\partial \mathcal{L_{\text {n}}}}{\partial \mathbf{p}_i}$ of position is independent of the normal $\mathbf{n}$. Consequently, the \textbf{supervision on rendered normal cannot effectively affect} the Gaussian position $\mathbf{p}$.
        
        \vspace{2mm}
        \noindent \textbf{{Case 2:}}
        From Eq.~\ref{eq:dd_dp}, there is an additional term with $\frac{\mathbf{n}}{\mathbf{n}\cdot\mathbf{p}}$, where the denominator $\mathbf{n} \cdot \mathbf{r}$ is a scalar term. This effectively makes the change in the position $ \frac{\partial \hat{D}}{\partial \mathbf{p}_i}$ to move along the direction of the normal $\mathbf{n}$. Consequently, the \textbf{supervision on D-Normal directly affects} the Gaussian position $\mathbf{p}$. 
        
        \vspace{2mm}
        \noindent We can further deduce that the gradient-update on the Gaussian position \textbf{pulls the position along the normal} towards the surface, which achieves better reconstruction. 
        

        In view of the above proof, we conclude that it is better to do supervision on the D-Normal regularizer. In addition to the mathematical proof, we also visualize the positions of Gaussian centers from the optimized 3D scenes both with and without the D-Normal regularizer 
        in Fig.~\ref{fig:wodn}, thereby providing experimental validation of the conclusion.

    \subsection{Implementation Details}
        We use PyTorch 2.0.1 and CUDA 11.8. for most experiments. All experiments are conducted on an NVIDIA 3090/4090/A5000/A6000 GPU. We set most hyperparameters to the same as that used in Gaussian Splatting~\cite{kerbl3Dgaussians}. For outdoor scenes in the TNT dataset, we also utilize decoupled appearance modeling~\cite{lin2024vastgaussian} to alleviate the exposure issue. Moreover, to remove some outlier Gaussians, we adopt a pruning technique from LightGaussian~\cite{fan2023lightgaussian}. We also use a cuboid bounding box to contain the scene we need to reconstruct and we only regulate and add the new densification to Gaussians inside the box. We use the same train and test data with 2DGS on TNT, Mip-NeRF360, and DTU datasets.

\begin{table}[htbp]
    \centering
    \caption{\textbf{Additional ablation on TNT Dataset.}}
    \begin{tabular}{lccc}
    \hline
     & Precision~$\uparrow$ & Recall~$\uparrow$ & F-score~$\uparrow$ \\
    \hline
    F. w/o semantic trimming & 0.37 & 0.42 & 0.38 \\
    G. w/o scale regularization & 0.35 & 0.38 & 0.36 \\
    \hline
    H. Set scale $s_3$ zero & 0.36 & 0.40 & 0.37 \\
    I. Scale Regularization & 0.39 & 0.42 & 0.40 \\
    \hline
    \end{tabular}
        \label{tab:additional_ablation}
\end{table}

    \subsection{Additional Ablation Studies}
        We further verify the effectiveness of additional design choices on reconstruction quality. We conduct experiments on the TNT dataset~\cite{knapitsch2017tanks} and report the F1-score. The quantitative result is reported in \tabref{tab:additional_ablation}. We first verify the proposed semantic trimming strategy (\tabref{tab:additional_ablation} F). From the table, we can see the strategy can increase the F1-score by 0.02. We can also observe the qualitative result from \figref{fig:semantic} that the trimming strategy can prune the sky correctly. Additionally, we confirm the effectiveness of scale regularization (\tabref{tab:additional_ablation} G), which leads to an improvement of 0.04. The current work~\cite{Huang2DGS2024} sets the last item of the scale factor to zero to flatten the 3D Gaussians for surface reconstruction. Additionally, we also ablate the setting zero (\tabref{tab:additional_ablation} H) and the scale regularization (\tabref{tab:additional_ablation} I). From the table, we can see that using the scale regularization instead of setting zero can obtain a 0.03 F1-score improvement.

        \begin{figure*}[h]
        	\centering
        	\includegraphics[width=\linewidth]{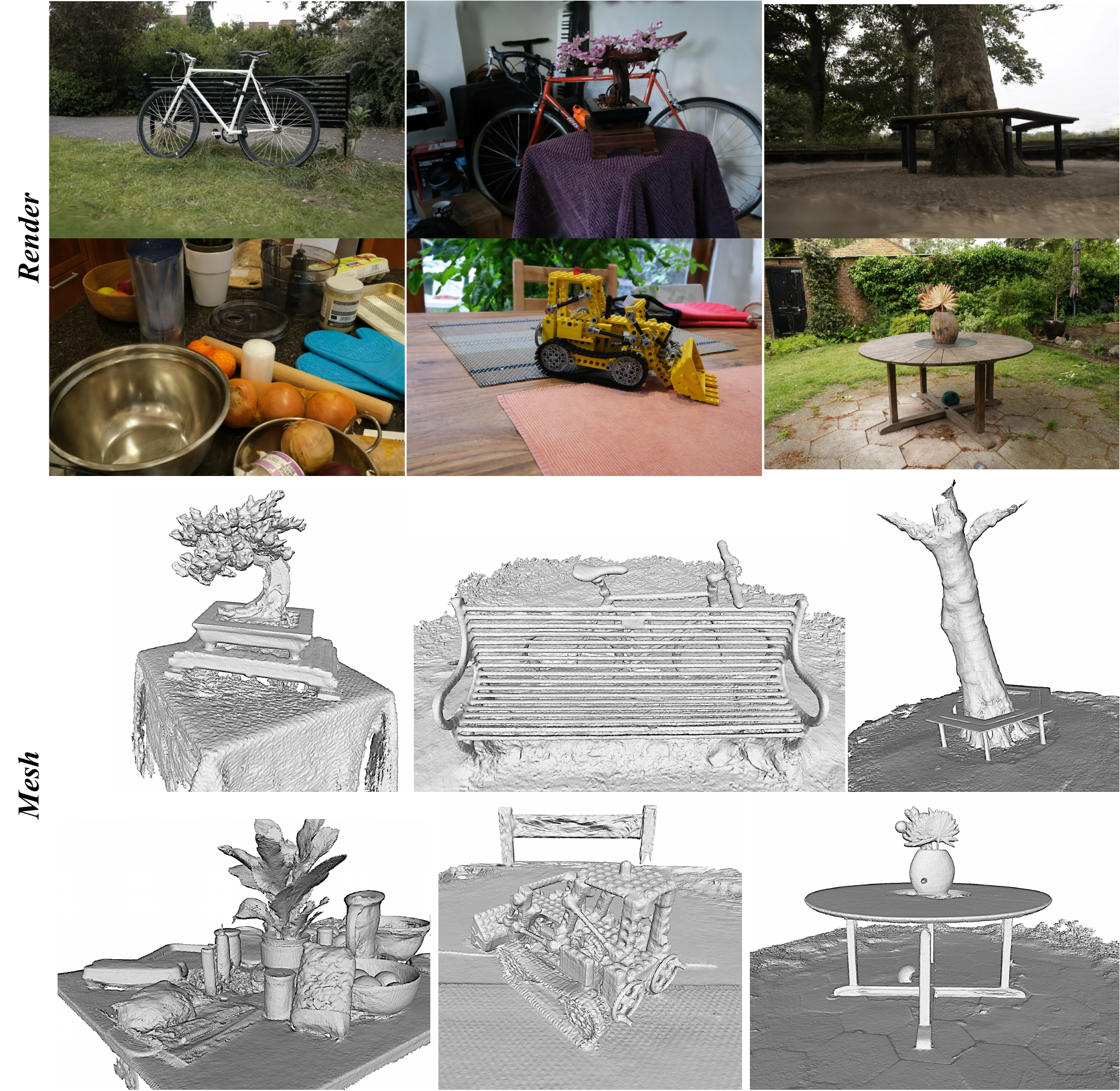} 
         \caption{\textbf{Qualitative results on the Mip-NeRF360 dataset.} Our method reconstructs surfaces with fine geometry details and produces high-fidelity renderings on Mip-NeRF360 dataset.
         }
         \label{fig:360}
        \end{figure*}
        \begin{figure*}[h]
        	\centering
        	\includegraphics[width=\linewidth]{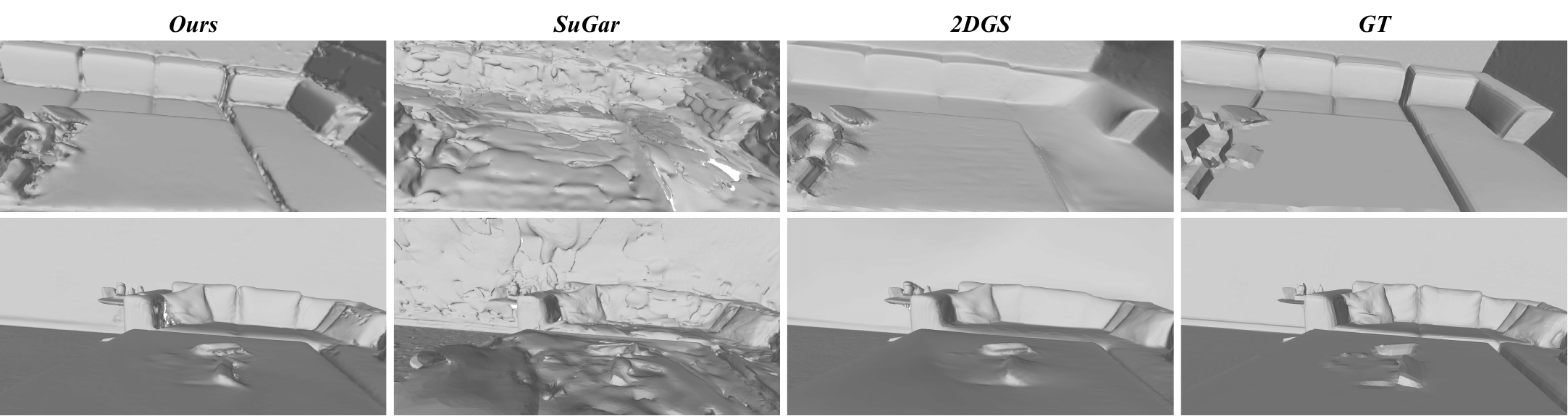} 
         \caption{\textbf{Qualitative results on the Replica dataset.} }
         \label{fig:replica_mesh}
        \end{figure*}
        
    \subsection{Additional Qualitative Results}
\begin{figure*}[t]
        	\centering
        	\includegraphics[width=\linewidth]{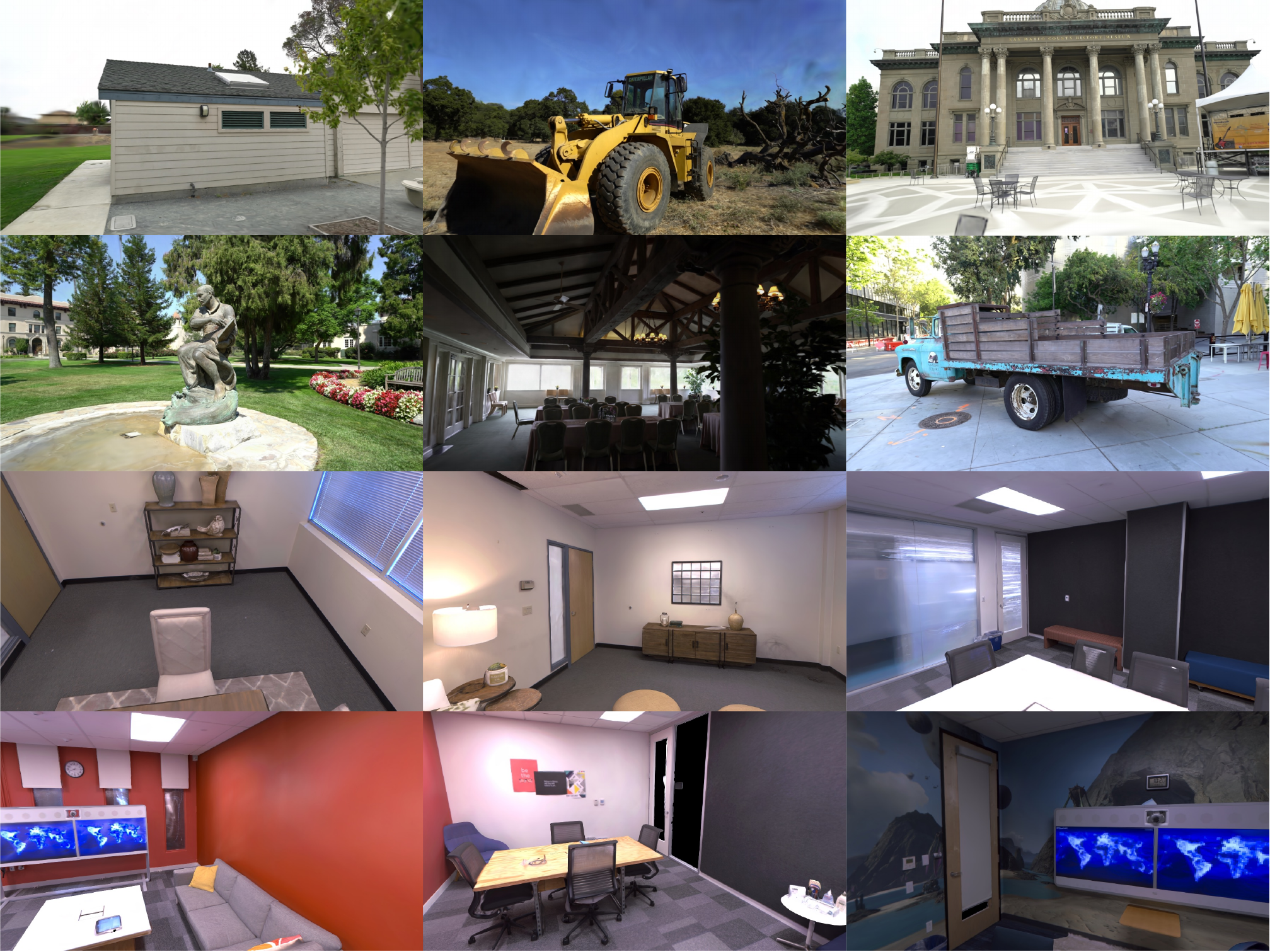}
         \caption{\textbf{Qualitative rendering results on the TNT and Replica dataset.}}
         \label{fig:tnt_replica_dataset}
        \end{figure*}

        \figref{fig:360} shows the rendering (top) and reconstruction (down) results on the Mip-NeRF360~\cite{barron2021mip} dataset. The rendering results on the TNT and Replica datasets are shown in \figref{fig:tnt_replica_dataset}. We also compare the qualitative results of VCR-GauS with SuGar and 2DGS, shown in \figref{fig:replica_mesh}. From the figure, we can see that our method can reconstruct both complete and high-detailed surfaces. In addition to the visualization in the form of pictures, we have also recorded a video in the supplementary material that can be downloaded and watched.
        
    \subsection{Additional Dataset}
        \setlength\tabcolsep{0.2em}
\begin{table*}[h]
    \centering
    \caption{\textbf{Quantitative comparison on the DTU Dataset~\cite{dtu}}. We show the Chamfer distance and average optimization time. 
    }
    \vspace{-0.2cm}
    \resizebox{.98\textwidth}{!}{
    \begin{tabular}{@{}llcccccccccccccccclcc}
    \toprule
     \multicolumn{3}{c}{Method}& 24 & 37 & 40 & 55 & 63 & 65 & 69 & 83 & 97 & 105 & 106 & 110 & 114 & 118 & 122 & & Mean & Time \\
     \midrule
    \multirow{4}{*}{\rotatebox[origin=c]{90}{Implicit}} & NeRF~\cite{nerf} & & 1.90 & 1.60 & 1.85 & 0.58 & 2.28 & 1.27 & 1.47 & 1.67 & 2.05 & 1.07 & 0.88 & 2.53 & 1.06 & 1.15 & 0.96 & & 1.49 & >12h \\
     & VolSDF~\cite{yariv2021volume} & &  1.14 & \tbest 1.26 &  0.81 & 0.49 & 1.25 &  \sbest0.70 &  \sbest0.72 &  \best 1.29 & \tbest 1.18 &  \sbest 0.70 & \tbest 0.66 & \tbest 1.08 &  0.42 &  \tbest 0.61 &  0.55 & & 0.86 & >12h \\
     & NeuS~\cite{wang2021neus} & &  1.00 & 1.37 & 0.93 & \sbest 0.43 &  1.10 &  \best 0.65 &   \best 0.57 &  1.48 &  \sbest 1.09 &  0.83 &  \sbest 0.52 &  1.20 & \best 0.35 &  \sbest 0.49 & \sbest 0.54 & &  0.84 & >12h \\
     & MonoSDF~\cite{yu2022monosdf} & & \tbest 0.83 & 1.61 & \tbest 0.65 & 0.47 & \best 0.92 & 0.87 & 0.87 & \sbest 1.30 & \tbest 1.25 & \best 0.68 & \tbest 0.65 & \sbest 0.96 & \tbest 0.41 & \tbest 0.62 & 0.58 & & 0.84 & >12h \\
     \midrule
    \multirow{4}{*}{\rotatebox[origin=c]{90}{Explicit}} 
    &  3DGS~\cite{kerbl3Dgaussians} & & 2.14 & 1.53 & 2.08 & 1.68 & 3.49 & 2.21 & 1.43 & 2.07 & 2.22 & 1.75 &  1.79 & 2.55 & 1.53 & 1.52 & 1.50 & & 1.96 & \ <~1h \\
     &  SuGaR~\cite{guedon2023sugar} & & 1.47 & 1.33 & 1.13 & 0.61 & 2.25 & 1.71 & 1.15 & 1.63 & 1.62 & 1.07 & 0.79 & 2.45 & 0.98 & 0.88 & 0.79 & & 1.33 & \ <~1h \\
     & 2DGS~\cite{Huang2DGS2024} &&  \best 0.48 &  \best0.91 &  \best 0.39 &  \best 0.39 &  \tbest 1.01 &  \tbest 0.83 &  \tbest 0.81 & \tbest 1.36 &  1.27 & \tbest 0.76  &  0.70 &  1.40 &   \sbest 0.40 &   0.76 &  \best 0.52 &&  \best 0.80 &  \ <~1h \\
     & Ours & & \sbest 0.55 & \best 0.91 & \sbest 0.40 & \sbest 0.43 & \sbest 0.97 & 0.95 & 0.84 & 1.39 & 1.30 & 0.90 & 0.76 & \best 0.92 & 0.44 & 0.75 & \sbest 0.54 & & \best 0.80 & \ <~1h \\
     \bottomrule
    \end{tabular}
    }
    \label{tab:dtu_result}
    \vspace{-0.1cm}
\end{table*}

\begin{figure*}[ht]
        	\centering
        	\includegraphics[width=\linewidth]{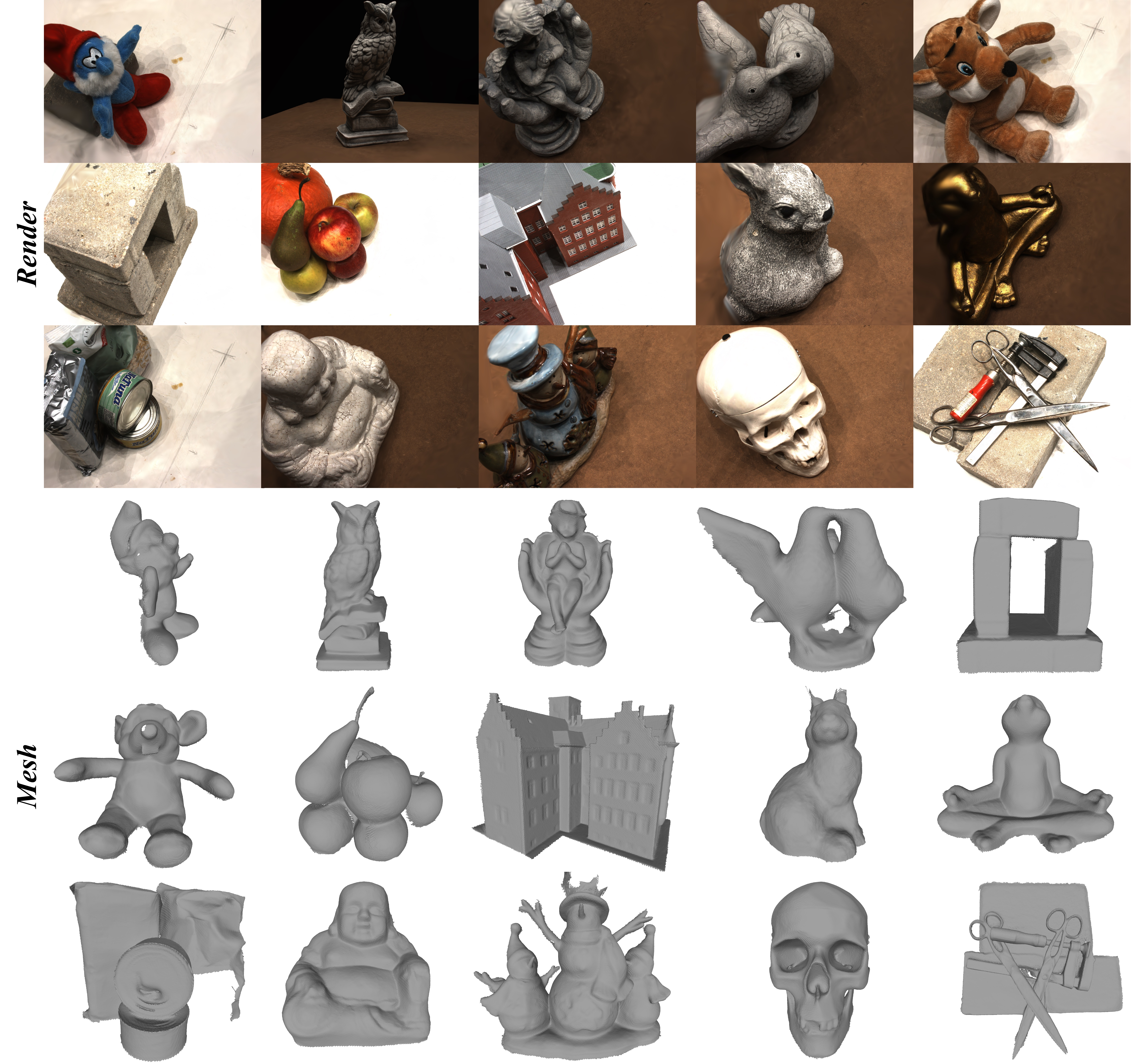} 
         \caption{\textbf{Qualitative results on the DTU dataset.}}
         \label{fig:dtu}
        \end{figure*}

        In this section, we report the result of object-level reconstruction. Although we focus on large-scale reconstruction, we still outperform implicit (\textit{i.e.}, NeRF, VolSDF, NeuS, and MonoSDF) and most explicit methods (\textit{i.e.}, 3DGS and SuGaR) on DTU~\cite{dtu}. While our method is comparable with current work 2DGS on object-level reconstruction, our method is much better than it on large-scale reconstruction, as shown in \tabref{tab:tnt} and \tabref{tab:replica}. The qualitative results on DTU are shown in \figref{fig:dtu}.

\section{Limitations}
\begin{figure*}[h]
        	\centering
        	\includegraphics[width=\linewidth]{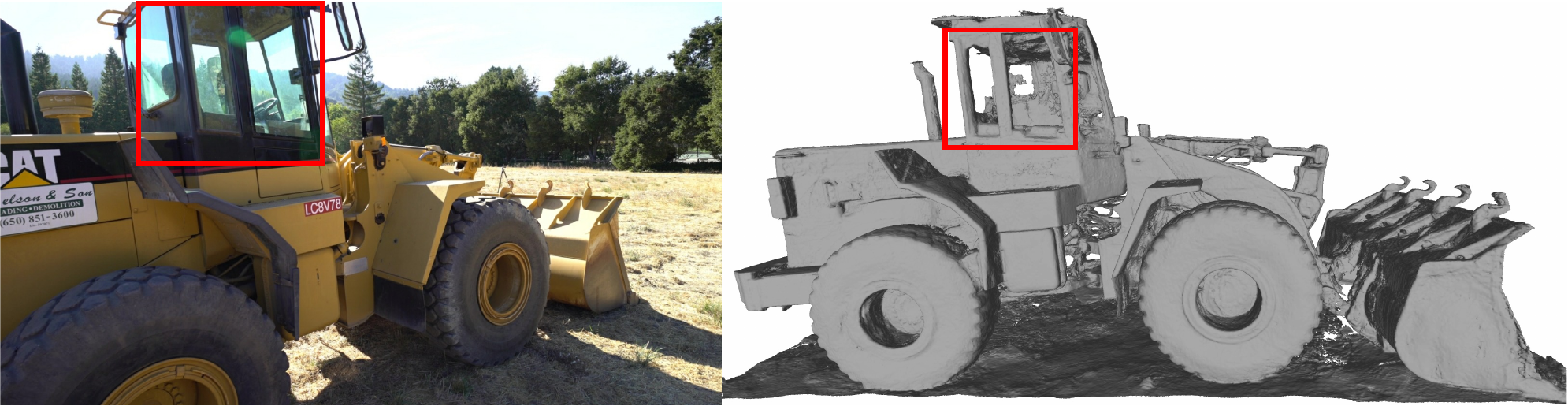} 
         \caption{\textbf{Illustration of limitations}: Our VCR-GauS encounters difficulties in accurately reconstructing semi-transparent surfaces, such as the window depicted in \textit{Caterpillar}.}
         \label{fig:failure}
        \end{figure*}

    Although our method can alleviate the problem of inaccurate normal prediction from a pretrained normal estimator, especially the inconsistent normal predictions across views, it fails under the extreme case when almost all predicted normal across views are wrong. In addition, our method cannot reconstruct beyond the observed scene. Furthermore, our method cannot capture the surface of the semi-transparent object as shown in \figref{fig:failure}.

\clearpage

    \section*{NeurIPS Paper Checklist}

\begin{enumerate}

\item {\bf Claims}
    \item[] Question: Do the main claims made in the abstract and introduction accurately reflect the paper's contributions and scope?
    \item[] Answer: \answerYes{} 
    \item[] Justification: We have discussed them on page 2.

\item {\bf Limitations}
    \item[] Question: Does the paper discuss the limitations of the work performed by the authors?
    \item[] Answer: \answerYes{} 
    \item[] Justification: We have discussed about that in the supplementary material.

\item {\bf Theory Assumptions and Proofs}
    \item[] Question: For each theoretical result, does the paper provide the full set of assumptions and a complete (and correct) proof?
    \item[] Answer: \answerNA{} 
    \item[] Justification: We do not have theoretical contributions in this work, where our contributions are validated with experiments.

    \item {\bf Experimental Result Reproducibility}
    \item[] Question: Does the paper fully disclose all the information needed to reproduce the main experimental results of the paper to the extent that it affects the main claims and/or conclusions of the paper (regardless of whether the code and data are provided or not)?
    \item[] Answer: \answerYes{} 
    \item[] Justification: All the hyper-parameters and network organizations are provided in the main paper and the appendix.

\item {\bf Open access to data and code}
    \item[] Question: Does the paper provide open access to the data and code, with sufficient instructions to faithfully reproduce the main experimental results, as described in supplemental material?
    \item[] Answer: \answerNo{} 
    \item[] Justification: Our codes have been released.

\item {\bf Experimental Setting/Details}
    \item[] Question: Does the paper specify all the training and test details (e.g., data splits, hyperparameters, how they were chosen, type of optimizer, etc.) necessary to understand the results?
    \item[] Answer: \answerYes{} 
    \item[] Justification: Training and test details are described in the main paper and the appendix.

\item {\bf Experiment Statistical Significance}
    \item[] Question: Does the paper report error bars suitably and correctly defined or other appropriate information about the statistical significance of the experiments?
    \item[] Answer: \answerNo{} 
    \item[] Justification: We follow existing related works for the setting of error bars.

\item {\bf Experiments Compute Resources}
    \item[] Question: For each experiment, does the paper provide sufficient information on the computer resources (type of compute workers, memory, time of execution) needed to reproduce the experiments?
    \item[] Answer: \answerYes{} 
    \item[] Justification: We have provided that in the supplementary material.
    
\item {\bf Code Of Ethics}
    \item[] Question: Does the research conducted in the paper conform, in every respect, with the NeurIPS Code of Ethics \url{https://neurips.cc/public/EthicsGuidelines}?
    \item[] Answer: \answerYes{} 
    \item[] Justification: We have reviewed that and claim we conform that Code of Ethics.

\item {\bf Broader Impacts}
    \item[] Question: Does the paper discuss both potential positive societal impacts and negative societal impacts of the work performed?
    \item[] Answer: \answerNA{} 
    \item[] Justification: Our method focuses on reconstructing 3D surfaces using Gaussian Splatting technique, which is a component of 3D reconstruction. It does not have further societal impacts than existing 3D reconstruction works.
    
\item {\bf Safeguards}
    \item[] Question: Does the paper describe safeguards that have been put in place for responsible release of data or models that have a high risk for misuse (e.g., pretrained language models, image generators, or scraped datasets)?
    \item[] Answer: \answerNA{} 
    \item[] Justification: The paper does not have such risks.

\item {\bf Licenses for existing assets}
    \item[] Question: Are the creators or original owners of assets (e.g., code, data, models), used in the paper, properly credited and are the license and terms of use explicitly mentioned and properly respected?
    \item[] Answer: \answerYes{} 
    \item[] Justification: We have cited them in the references.

\item {\bf New Assets}
    \item[] Question: Are new assets introduced in the paper well documented and is the documentation provided alongside the assets?
    \item[] Answer: \answerNA{} 
    \item[] Justification: We use and cite existing datasets in this work. Other assets including code/model will be released after submitting.

\item {\bf Crowdsourcing and Research with Human Subjects}
    \item[] Question: For crowdsourcing experiments and research with human subjects, does the paper include the full text of instructions given to participants and screenshots, if applicable, as well as details about compensation (if any)? 
    \item[] Answer: \answerNA{} 
    \item[] Justification: We do not include such experiments.

\item {\bf Institutional Review Board (IRB) Approvals or Equivalent for Research with Human Subjects}
    \item[] Question: Does the paper describe potential risks incurred by study participants, whether such risks were disclosed to the subjects, and whether Institutional Review Board (IRB) approvals (or an equivalent approval/review based on the requirements of your country or institution) were obtained?
    \item[] Answer: \answerNA{} 
    \item[] Justification: We do not include such experiments.

\end{enumerate}

\end{document}